\documentclass[10pt,twocolumn,letterpaper]{article}

\usepackage[pagenumbers]{iccv} 

\usepackage{balance}
\usepackage{xcolor}
\usepackage{makecell}
\usepackage{tabularx}
\newcolumntype{Y}{>{\centering\arraybackslash}X}
\definecolor{iccvblue}{rgb}{0.21,0.49,0.74}
\usepackage[pagebackref,breaklinks,colorlinks,allcolors=iccvblue]{hyperref}

\def\paperID{1789} 
\def\confName{ICCV}
\def\confYear{2025}

\title{\vspace{-0.5cm}InfiniteYou: Flexible Photo Recrafting While Preserving Your Identity\vspace{-0.3cm}}

\author{Liming Jiang \hspace{13pt} Qing Yan \hspace{13pt} Yumin Jia \hspace{13pt} Zichuan Liu \hspace{13pt} Hao Kang \hspace{13pt} Xin Lu
\\[4pt]
ByteDance Intelligent Creation\\[4pt]
{\normalsize Project Page: \href{https://bytedance.github.io/InfiniteYou}{\bf\textcolor{magenta}{https://bytedance.github.io/InfiniteYou}}}
}

\begin{document}


\twocolumn[{
\renewcommand\twocolumn[1][]{#1}
\maketitle

\begin{center}
    \centering
    \vspace{-0.6cm}
    \includegraphics[width=0.98\linewidth]{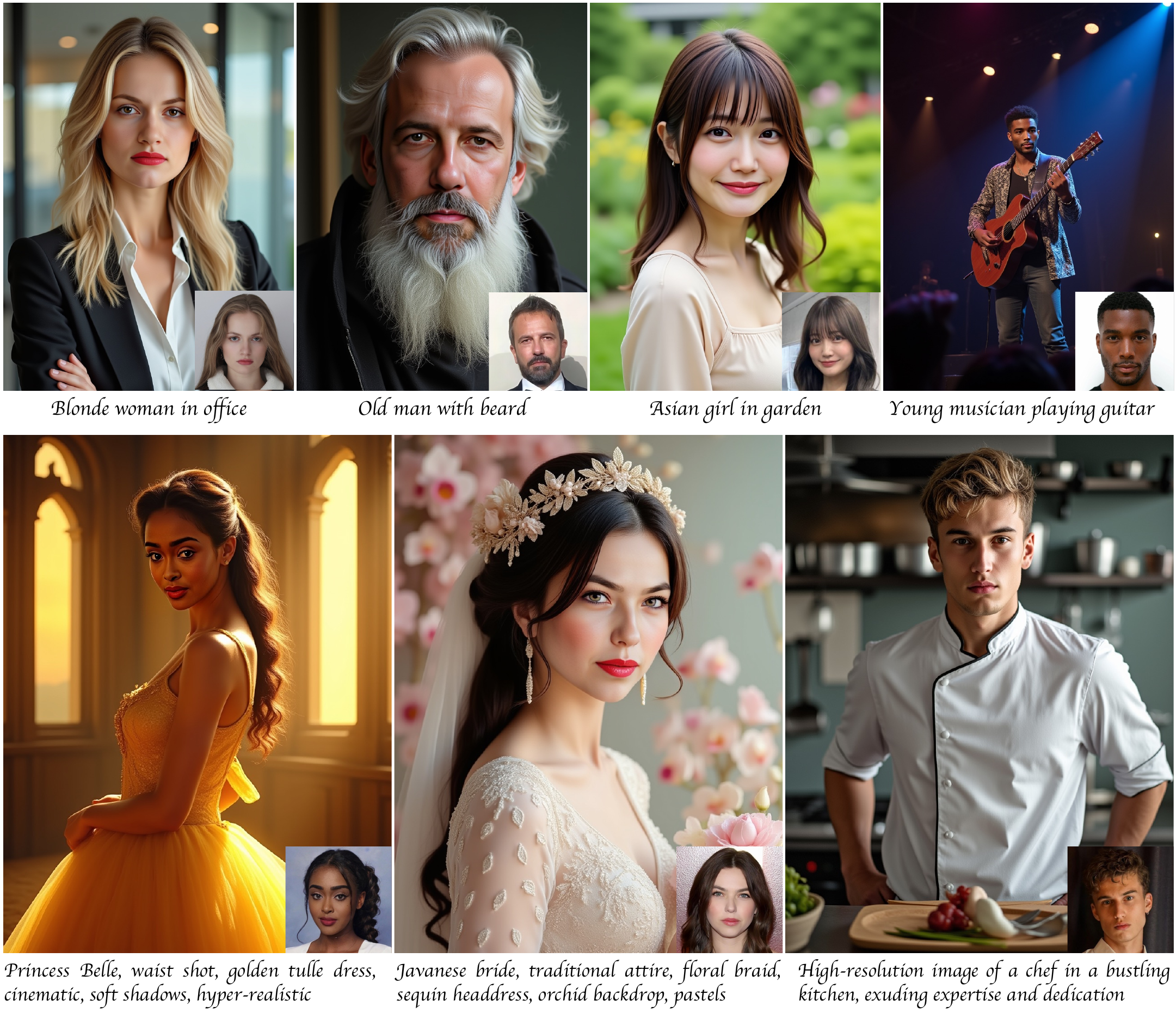}
    \vspace{-0.2cm}
    \captionof{figure}{InfiniteYou generates identity-preserved images with exceptional identity similarity, text-image alignment, quality, and aesthetics.}
    \vspace{0.11cm}
    \label{fig:teaser}
\end{center}
}]

\begin{abstract}
\label{sec:abstract}
Achieving flexible and high-fidelity identity-preserved image generation remains formidable, particularly with advanced Diffusion Transformers (DiTs) like FLUX. We introduce \textbf{InfiniteYou (InfU)}, one of the earliest robust frameworks leveraging DiTs for this task. InfU addresses significant issues of existing methods, such as insufficient identity similarity, poor text-image alignment, and low generation quality and aesthetics. Central to InfU is InfuseNet, a component that injects identity features into the DiT base model via residual connections, enhancing identity similarity while maintaining generation capabilities. A multi-stage training strategy, including pretraining and supervised fine-tuning (SFT) with synthetic single-person-multiple-sample (SPMS) data, further improves text-image alignment, ameliorates image quality, and alleviates face copy-pasting. Extensive experiments demonstrate that InfU achieves state-of-the-art performance, surpassing existing baselines. In addition, the plug-and-play design of InfU ensures compatibility with various existing methods, offering a valuable contribution to the broader community. Code and model: \href{https://github.com/bytedance/InfiniteYou}{\textcolor{magenta}{https://github.com/bytedance/InfiniteYou}}. 
\end{abstract}    
\section{Introduction}
\label{sec:intro}

Identity-preserved image generation aims to recraft a photograph of a specific person using free-form text descriptions while preserving facial identity. This task is challenging but highly beneficial.
Previous methods~\cite{ipa, instantid, pulid} have been mainly developed on U-Net~\cite{unet}-based text-to-image diffusion models~\cite{diffusion, DDPM, LDM}, such as Stable Diffusion XL~(SDXL)~\cite{sdxl}. However, due to the limited generation capacity of the base model, the quality of generated images remains inadequate.
Recent strides in Diffusion Transformers (DiTs)~\cite{dit, sd3} have made remarkable progress in content creation.
In particular, the latest releases of state-of-the-art rectified flow DiTs, such as FLUX~\cite{flux} and Stable Diffusion 3.5~\cite{sd3}, showcase incredible image generation quality.
Consequently, it is crucial to explore solutions that can leverage the immense potential of DiTs for downstream applications like identity-preserved image generation.

DiT-based identity-preserved image generation remains formidable due to several factors: the absence of customized module designs, difficulties in model scaling, and a lack of high-quality data. Thus, effective solutions for this task using state-of-the-art rectified flow~\cite{rfliu,rfalbergo} DiTs, such as FLUX~\cite{flux}, are currently scarce in both academia and industry.
PuLID-FLUX, derived from PuLID~\cite{pulid}, presented an initial attempt to develop an identity-preserved image generation model based on FLUX.
Other open-source efforts, including FLUX.1-dev IP-Adapters from InstantX~\cite{instantxfluxipa} and XLabs-AI~\cite{xfluxipa}, are not tailored for human facial identities.
Nevertheless, existing methods fall short in three key aspects: \textbf{1)} The identity similarity is not sufficient; \textbf{2)} The text-image alignment and editability are poor, and the face copy-paste issue is evident; \textbf{3)} The generation capability of FLUX is largely compromised, resulting in lower image quality and aesthetic appeal.

To address the aforementioned challenges, we propose a simple yet effective framework for identity-preserved image generation, namely InfiniteYou (InfU). This framework is designed to be systematic and robust, enabling flexible text-based photo re-creation for diverse identities, races, and age groups across various scenarios. 
We introduce InfuseNet, a generalization of ControlNet~\cite{controlnet}, which ingests identity information along with the controlling conditions.
The projected identity features are injected by InfuseNet into the DiT base model through residual connections, thereby disentangling text and identity injections.
InfuseNet is both scalable and compatible, harnessing the powerful generation capabilities of DiTs.
The scale-up injection network and the delicate architecture design, equipped with large-scale model training, effectively enhance identity similarity.
To improve text-image alignment, image quality, and aesthetic appeal, we employ a multi-stage training strategy, including pretraining and supervised fine-tuning~(SFT)~\cite{jiang2024supervised,touvron2023llama}. The SFT stage utilizes carefully designed synthetic single-person-multiple-sample~(SPMS) data generation, leveraging our pretrained model itself and various off-the-shelf modules.
This strategy enhances the quantity, quality, aesthetics, and text-image alignment of the training data, thus improving overall model performance and alleviating the face copy-paste issue.
Thanks to the InfuseNet design, identity information is injected purely via residual connections between DiT blocks, unlike conventional practices~\cite{ipa, instantid, pulid} that directly modify attention~\cite{attention} layers via IP-Adapter~(IPA).
Consequently, the generation capability of the base model remains largely intact, allowing for the generation of high-quality and aesthetically pleasing images.
Moreover, InfU is plug-and-play and readily compatible with many other methods or plugins, thus offering significant value to the broader community.

Comprehensive experiments show that the proposed InfU framework achieves state-of-the-art performance 
(see Figure~\ref{fig:teaser}), significantly surpassing existing baselines in identity similarity, text-image alignment, and overall image quality. Our main contributions are summarized as follows:
\begin{itemize}
    \item We propose InfiniteYou (InfU), a versatile and robust DiT-based framework for flexible identity-preserved image generation under various scenarios.
    \item We introduce InfuseNet, a generalization of ControlNet, which effectively injects identity features into the DiT base model via residual connections, enhancing identity similarity with minimal impact on generation capabilities.
    \item We employ a multi-stage training strategy, including pretraining and supervised fine-tuning (SFT), using synthetic single-person-multiple-sample (SPMS) data generation. This approach significantly improves text-image alignment, image quality, and aesthetic appeal. 
    \item The InfU module features a desirable plug-and-play design, compatible with many existing methods, thus providing a valuable contribution to the broader community.
\end{itemize}

\section{Related Work}
\label{sec:related}

\noindent
\textbf{Text-to-image Diffusion Transformers (DiTs).} 
Diffusion models~\cite{diffusion,DDPM,DDIM,LDM} have become a standard paradigm given their incredible capability to produce high-fidelity images. Text-to-image generation aims to synthesize images through the denoising diffusion process~\cite{DDPM, LDM} from a Gaussian distribution given textual descriptions. 
Conventional methods~\cite{LDM, sdxl, advdistill, sdxllightning} are based on U-Net~\cite{unet}, where the text representation is extracted by CLIP~\cite{clip}.
Recent DiTs~\cite{dit}, based on Vision Transformers (ViTs)~\cite{vit,scalingvit} and typically using text embeddings encoded by T5~\cite{t5} in addition to CLIP, have exhibited even higher generation capacity compared to U-Net architectures.
The latest releases of DiTs with rectified flows (RFs)~\cite{rfliu,rfalbergo}, such as Stable Diffusion~3.5~\cite{sd3}, Playground~V3~\cite{playgroundv3}, Recraft~V3~\cite{recraftv3}, and FLUX.1~\cite{flux}, have further shown their unprecedented generation quality.
The progress made by these DiTs naturally stimulates the development of customized approaches in their downstream applications. This highlights the significance of our work, reforming the architecture from U-Nets to DiTs for identity-preserved image generation.

\noindent
\textbf{Identity-preserved image generation.}
Tuning-based methods for identity-preserved image generation include~\cite{textualinv,dreambooth,lora,hyperdreambooth,customdiffusion}. However, their practical significance is hindered by low efficiency and high tuning cost due to their specificity to individual identities.
Therefore, tuning-free methods have become the mainstream practice for this task. A series of efforts, such as IP-Adapter~\cite{ipa}, FastComposer~\cite{fastcomposer}, Photomaker~\cite{photomaker}, InstantID~\cite{instantid}, FlashFace~\cite{flashface}, Arc2Face~\cite{arc2face}, Imagine Yourself~\cite{imagineyourself}, and PuLID~\cite{pulid}, typically employ additional trainable modules as adapters to inject identity information. 
After training, these approaches can generate customized images without additional tuning, even when ingesting new subject samples. However, these methods are mainly developed for U-Net-based Stable Diffusion~\cite{LDM} and SDXL~\cite{sdxl}, where the limited generative capability of the base model inevitably constrains the quality of the generated images.
The remarkable achievements of DiTs highlight the importance of base model replacement, which remains underexplored.
PuLID-FLUX~\cite{pulid} made an initial attempt based on IP-Adapter trained with alignment and identity losses~\cite{pulid}. Other open-source efforts, including FLUX.1-dev IP-Adapters from InstantX~\cite{instantxfluxipa} and XLabs-AI~\cite{xfluxipa}, were devised but not tailored for human faces.
Nevertheless, existing methods still face limitations: insufficient identity similarity, poor text-image alignment, face copy-paste issues, and compromised generation quality.
The proposed InfU effectively addresses these shortcomings.

\section{Methodology}
\label{sec:method}

\begin{figure}
	\centering
	\includegraphics[width=\linewidth]{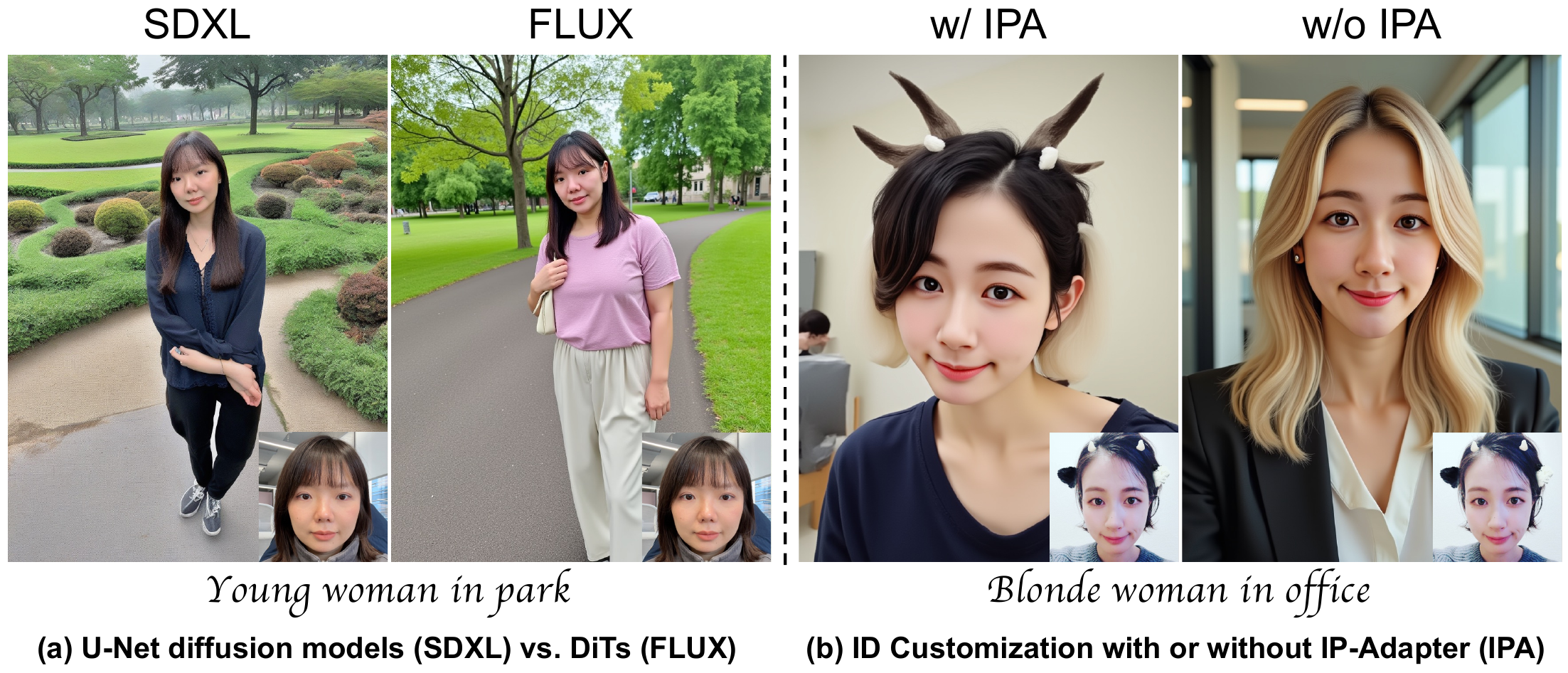}
	\vspace{-0.58cm}
	\caption{The superiority of the DiT-based method over the U-Net-based one and the side effects of IP-Adapter~(IPA)~\cite{ipa}.}
	\label{fig:inspiration}
	\vspace{-0.36cm}
\end{figure}

\subsection{Preliminary}
\label{subsec:method_prelimilary}

\noindent
\textbf{Simulation-free training of flows.}
Following~\cite{sd3}, generative models are defined to establish a transformation between samples $x_1$ drawn from a noise distribution $p_1$ to samples $x_0$ drawn from a data distribution $p_0$, formulated through an ordinary differential equation (ODE),
\begin{equation}
  \label{eq:ode}
  dy_t = v_\Theta\left(y_t, t\right)\,dt,
\end{equation}
where the velocity $v$ is parameterized by the neural network weights $\Theta$. The previous work~\cite{neuralode} proposed solving \cref{eq:ode} directly using differentiable ODE solvers.
However, this method is computationally intensive, particularly for large neural network structures that parameterize $v_\Theta\left(y_t, t\right)$. 
A more efficient approach is to directly regress a vector field $u_t$ that defines a probability path~\cite{flowmatching} between $p_0$ and $p_1$.
To formulate such a vector field $u_t$, a forward process is defined that corresponds to a probability path $p_t$ between $p_0$ and $p_1=\mathcal{N}\left(0, 1\right)$, expressed as
\begin{equation}
  \label{eq:forwardprocess}
  z_t = a_t x_0 + b_t \epsilon, \quad\text{where}\; \epsilon \sim \mathcal{N}(0,I).
\end{equation}
For $a_0 = 1, b_0 = 0, a_1 = 0$, and $b_1 = 1$, the marginals
\begin{align}
  \label{eq:marginals}
  p_t\left(z_t\right) &=
  \mathbb{E}_{\epsilon \sim \mathcal{N}\left(0,I\right)}
  p_t\left(z_t \vert \epsilon\right),
\end{align}
align with the data and noise distributions. A marginal vector field $u_t$ can generate the marginal probability paths $p_t$ using conditional vector fields $u_t\left(\cdot | \epsilon\right)$:
\begin{align}
    u_t\left(z\right) = \mathbb{E}_{\epsilon \sim \mathcal{N}\left(0,I\right)} u_t\left(z \vert \epsilon\right) \frac{p_t\left(z \vert \epsilon\right)}{p_t\left(z\right)}.
  \label{eq:marginal_u}
\end{align}
It is intractable to regress $u_t$ directly due to the marginalization in~\cref{eq:marginal_u}. Thus, we switch to a simple and tractable objective, \ie, Conditional Flow Matching~\cite{flowmatching,sd3}:
\begin{align}
   \label{eq:condflowmatch}
   \mathcal{L}_{CFM} =  \mathbb{E}_{t, p_t\left(z \vert \epsilon\right), p\left(\epsilon\right) } \left\| v_{\Theta}\left(z, t\right) - u_t\left(z \vert \epsilon\right) \right\|_2^2.
\end{align}

\noindent
\textbf{Rectified Flow.}
Rectified flows (RFs)~\cite{rfliu, rfalbergo} define the forward process as straight paths between the data distribution and a standard Gaussian distribution, \ie,
\begin{equation}
z_t = \left(1-t\right) x_0 + t \epsilon,
\end{equation}
where \( \epsilon \sim \mathcal{N}\left(0,I\right) \). The network output directly parameterizes the velocity \( v_\Theta \).
We use $\mathcal{L}_{CFM}$ (\cref{eq:condflowmatch}) as the loss objective.
Different flow trajectories and samplers are defined in~\cite{sd3}, including logit-normal sampling, which we also employ in our model training.

\begin{figure*}
	\centering
	\vspace{-0.55cm}
	\includegraphics[width=0.9\linewidth]{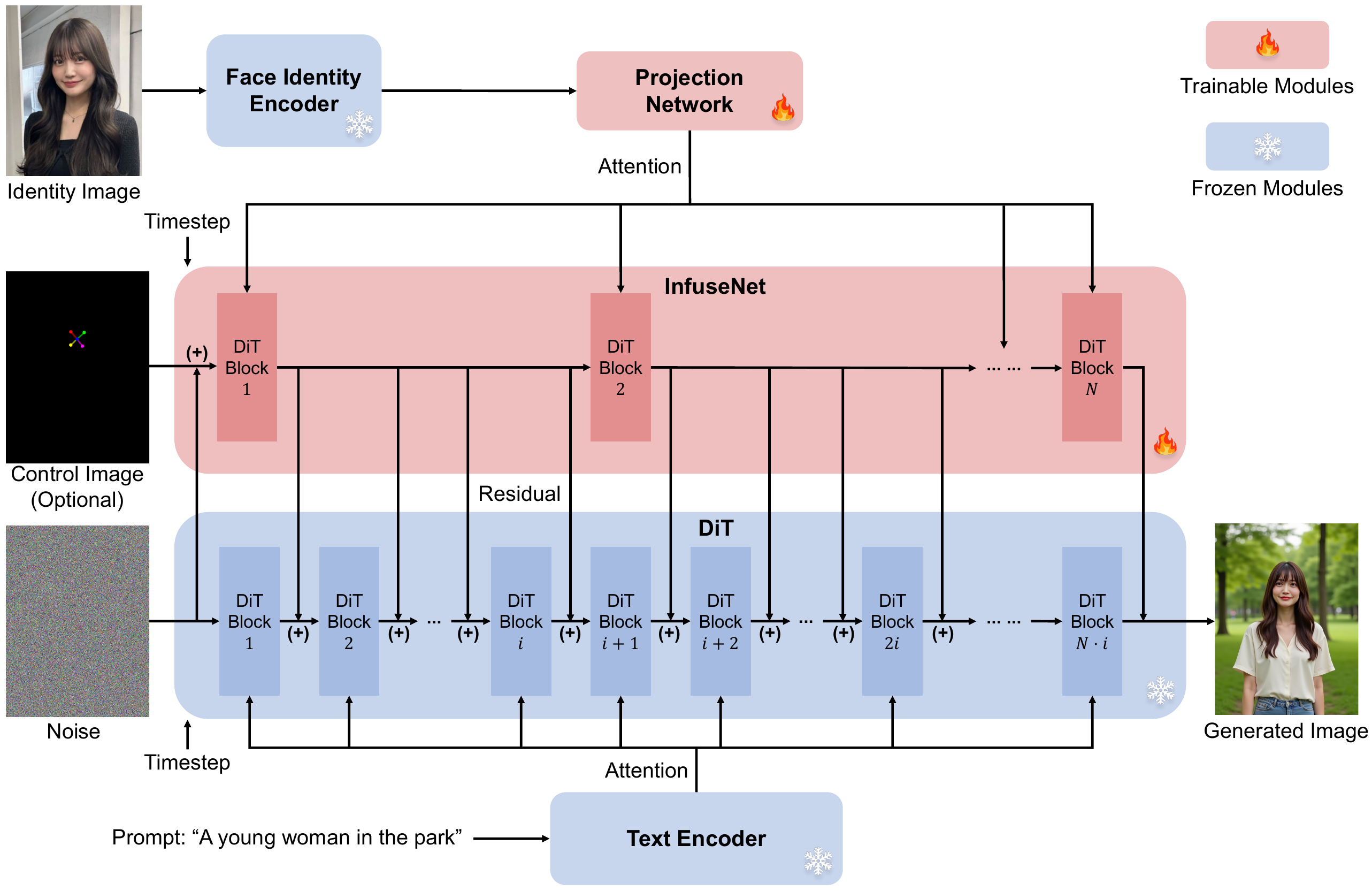}
	\vspace{-0.25cm}
	\caption{The main framework of InfiniteYou (InfU) and the detailed architecture of InfuseNet. The projected identity features and an optional control image are injected by InfuseNet into text-to-image DiTs via residual connections. Specifically, each DiT block in InfuseNet predicts the output residuals of the corresponding $i$ DiT blocks in the base model. Only InfuseNet and the projection network are trainable.}
	\label{fig:framework}
	\vspace{-0.4cm}
\end{figure*}

\noindent
\textbf{Text-to-image DiTs.}
Our general setup follows Stable Diffusion 3.5~\cite{sd3} and FLUX~\cite{flux}, derived from Latent Diffusion Models (LDM)~\cite{LDM} for training text-to-image models in the latent space of a pretrained autoencoder.
Apart from encoding images into latent representations, we also encode the text conditioning $c_\mathrm{text}$ using pretrained, frozen text models.
We use FLUX~\cite{flux} as our DiT base model, which uses T5-XXL~\cite{t5} and CLIP~\cite{clip} for text encoding.
FLUX uses a multimodal diffusion backbone, \ie, MMDiT~\cite{sd3}. Unlike traditional DiTs~\cite{dit}, MMDiT uses two separate sets of weights for the two modalities, given that text and image embeddings are conceptually different.
This setup is equivalent to having two independent Transformers for each modality, but combines the sequences via joint attention to ensure that both representations work in their own space while considering each other.
FLUX also applies several single DiT blocks~\cite{scalingvit} after MMDiT blocks.

In addition to text-conditional image generation, the proposed InfU method also injects human facial identity information $c_\mathrm{id}$ to accommodate additional modalities.

\begin{figure}
	\centering
	\vspace{-0.35cm}
	\includegraphics[width=0.99\linewidth]{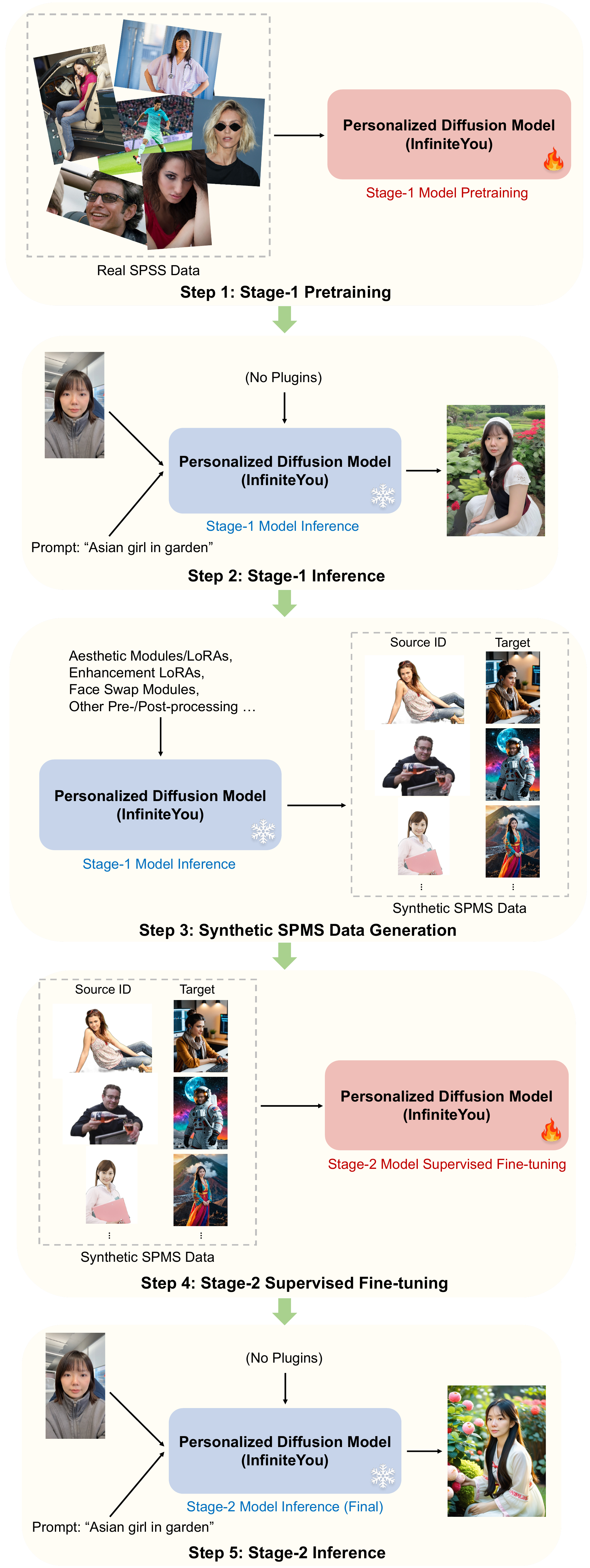}
	\caption{The introduced multi-stage training strategy with synthetic single-person-multiple-sample (SPMS) data and supervised fine-tuning (SFT).}
	\label{fig:mstraining}
\end{figure}

\subsection{Network Architecture}
\label{subsec:method_infusenet}

Conventional approaches~\cite{ipa, instantid} for this task were primarily developed for U-Net-based diffusion models like SDXL~\cite{sdxl}. However, the image quality generated by these methods remains inadequate (see Figure~\ref{fig:inspiration}~(a)). 
The significantly better performance of FLUX than SDXL in background clarity, human topology, small face quality, and overall appeal pinpoints the importance of DiT-based solutions. The proposed InfU is inspired by these efforts while presenting a novel solution built on DiTs. We focus on the development and comparison of DiT-based approaches due to their evident superiority over SDXL, as demonstrated.

Unlike common practices~\cite{ipa, instantid, pulid} that modify attention~\cite{attention} layers via IP-Adapter (IPA)~\cite{ipa} to inject identity information, we observe the non-optimality of IPA and avoid using it. As shown in Figure~\ref{fig:inspiration}~(b), IPA typically introduces side effects, such as degradation in text-image alignment, image quality and aesthetics. 
We deduce that directly modifying the attention layers significantly compromises the generative capability of the base model. In addition, injecting text and identity information at the same positions (\ie, attention layers) may bring potential entanglement and conflict, thus harming overall performance. Therefore, we propose a novel alternative solution without IPA, mitigating these issues while maintaining high identity similarity.

The proposed InfU framework is illustrated in Figure~\ref{fig:framework}. The DiT base model (\eg, FLUX) remains frozen during training and serves as the main branch for image generation. It ingests a noise map sampled from a standard Gaussian distribution, along with features from both the identity image and text prompt inputs, to generate an image that adheres to the text description while preserving the human facial identity through multiple denoising steps. 
The text prompt is embedded by a frozen text encoder and then fed into the base model through attention layers~\cite{sd3}. Below, we detail our mechanism for injecting identity information.

We introduce InfuseNet, an important branch that injects identity and control signals (see Figure~\ref{fig:framework}).
InfuseNet shares a similar structure with the DiT base model but contains fewer Transformer blocks. 
We denote $M$ as the number of DiT blocks in the base model and $N$ as the number of DiT blocks in InfuseNet. We have $M=N \cdot i$, where $i$ is the multiplication factor.  
An optional control image, such as a five-facial-keypoint image, can be input into InfuseNet to control the generation position of the subject. If no control is needed, a pure black image can be used instead.
The identity image is encoded by a frozen face identity encoder into identity embeddings, which are fed into a projection network. This network projects the identity features and sends them to InfuseNet through attention layers, similar to how text features are handled in the DiT base model. InfuseNet then predicts the output residual connections of the DiT base model, contributing to the final image synthesis. Specifically, DiT block $j$ in InfuseNet predicts the residuals of the following DiT blocks in the base model:
\begin{equation}
  \label{eq:infusenetresidual}
  \left(j-1\right) \cdot i + 1,\;\;\left(j-1\right) \cdot i + 2,\;\;\ldots,\;\;j \cdot i.
\end{equation}

During training, the projection network and InfuseNet are trainable (using $\mathcal{L}_{CFM}$ in~\cref{eq:condflowmatch}), while other modules remain frozen. The proposed InfuseNet can be seen as a generalization of ControlNet~\cite{controlnet}, capable of ingesting more modalities to influence the generation process via residual connections. This residual injection of identity features is distinct from the text injection through attention layers, which effectively separates text and identity inputs, thereby reducing potential entanglement and conflict.
Thanks to this pure residual injection design that does not rely on IPA, the generative capability of the base model is less compromised, resulting in higher quality and improved text-image alignment.
InfuseNet is also based on DiT, and its similar architecture with the base model ensures scalability and compatibility.
The scalable network design and large-scale training enhance identity similarity.

\subsection{Multi-Stage Training Strategy}
\label{subsec:method_mstraining}

Despite the robust network design of InfU, challenges in text-image alignment, generation aesthetics, and image quality degradation remain, especially in certain hard cases. This issue is critical for state-of-the-art approaches, necessitating a general solution to facilitate future research.

We devise a multi-stage training strategy, including pretraining and supervised fine-tuning~(SFT)~\cite{jiang2024supervised,touvron2023llama}.
This strategy improves the quantity, quality, aesthetics, and text-image alignment of training data, thereby enhancing overall model performance \wrt the above problems. The training strategy is formulated in the following steps (see Figure~\ref{fig:mstraining}).

\noindent
\textbf{Step 1:} We collect and filter real single-person-single-sample (SPSS) data from several human portrait datasets. The data, though not highly aesthetic or high-quality, can be used for stage-1 pretraining of our InfU model, following standard training practices~\cite{ipa, instantid}. Using real SPSS data, we employ a single authentic portrait image as both the source identity image and the generation target image to learn reconstruction during training.

\noindent
\textbf{Step 2:} After stage-1 pretraining of the InfU model, we conduct stage-1 model inference to evaluate image generation performance without any plugins, such as LoRA~\cite{lora}. While the facial identity similarity of the generated results is satisfactory, there remains room for improvement in text-image alignment, generation aesthetics, and image quality. 

\noindent
\textbf{Step 3:} We then equip the stage-1 trained InfU model with a series of useful off-the-shelf modules, such as aesthetic modules/LoRAs, enhancement LoRAs, face swap modules~\cite{simswap}, and other pre-/post-processing tools, \etc. Although time-consuming and cumbersome, this process enables the model to generate synthetic data with much higher quality and aesthetics. We intentionally formulate the data as single-person-multiple-sample (SPMS), where a real face image serves as the source identity image and the synthetic data serves as the generation target image. 

\noindent
\textbf{Step 4:} The synthetic SPMS data is subsequently fed into the stage-1 trained InfU model for stage-2 supervised fine-tuning (SFT). Leveraging the properties of SPMS, we use real face data as the source identity and the paired high-quality synthetic data as the generation target for model training. Other training settings remain similar to those in stage 1. This SFT enables the model to learn the high quality and aesthetics of the synthetic data while retaining identity similarity with the real face input.

\noindent
\textbf{Step 5:} After stage-2 SFT, the InfU model is ready for final inference and deployment. Without any plugins, the text-image alignment, generation aesthetics, and image quality of the generated results are significantly improved, while maintaining high facial identity similarity.

\begin{figure*}[t]
	\centering
	\vspace{-0.48cm}
	\includegraphics[width=0.99\linewidth]{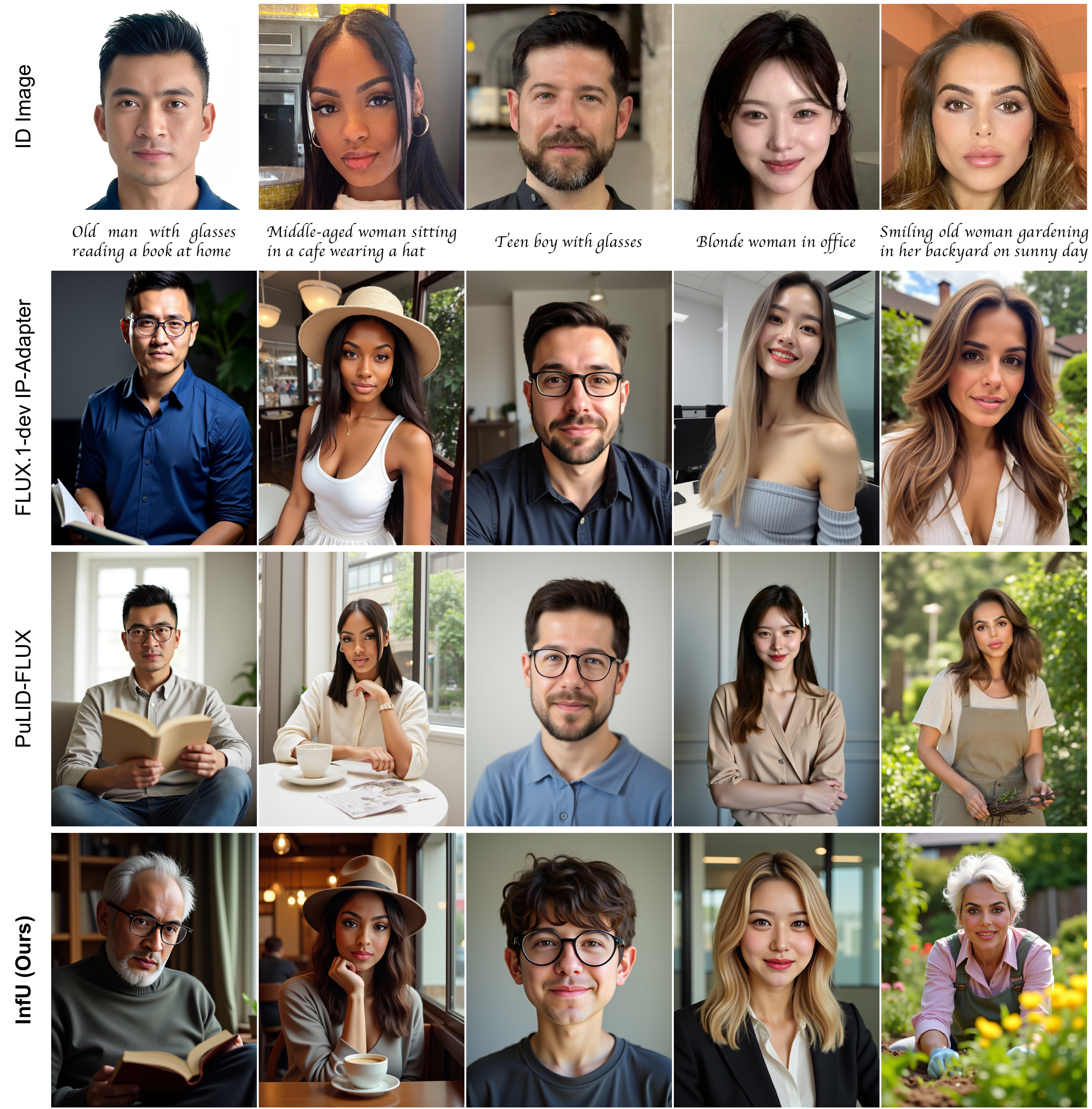}
	\vspace{-0.2cm}
	\caption{Qualitative comparison results of InfU with the state-of-the-art baselines, FLUX.1-dev IP-Adapter~\cite{instantxfluxipa} and PuLID-FLUX~\cite{pulid}.}
	\label{fig:qualitative_comparison}
	\vspace{-0.35cm}
\end{figure*}

\section{Experiments}
\label{sec:exp}

\subsection{Settings}
\label{subsec:exp_settings}

\noindent
\textbf{Implementation details.}
We implement our InfiniteYou (InfU) framework using PyTorch and leverage the Hugging Face Diffusers library. 
The DiT base model is FLUX.1-dev~\cite{flux}. 
We set the multiplication factor $i=4$ for InfuseNet. 
The projection network is derived from~\cite{ipa}, with the token number of the projected identity feature set to~$8$. 
All experiments are conducted using FSDP~\cite{fsdp} on NVIDIA H100 GPUs, each with $80$GB VRAM. We use the AdamW~\cite{adamw} optimizer with $\beta_1=0.9$ and $\beta_2=0.999$. The weight decay is set to $0.01$.
We employ Conditional Flow Matching~\cite{flowmatching,sd3} (\cref{eq:condflowmatch}) as the loss function with logit-normal sampling~\cite{sd3} of \texttt{rf/lognorm(0.00, 1.00)}. 
For stage-1 pretraining, the model is trained using an initial learning rate of $2 \times 10^{-5}$ on $128$ GPUs. The total batch size is set to $512$, and stage-1 training spans $300$k iterations.
For stage-2 supervised fine-tuning, the model is trained with an initial learning rate of $1 \times 10^{-5}$ on $64$ GPUs, with a total batch size of $256$. All other settings remain unchanged.

\noindent
\textbf{Datasets.}
For stage-1 pretraining, we use a total of nine open source datasets, including VGGFace2~\cite{cao2018vggface2}, MillionCelebs~\cite{zhang2020global}, CelebA~\cite{liu2015faceattributes}, CelebV-HQ~\cite{zhu2022celebvhq}, FFHQ~\cite{karras2019style}, VFHQ~\cite{xie2022vfhq}, EasyPortrait~\cite{EasyPortrait}, CelebV-Text~\cite{yu2022celebvtext}, CosmicManHQ-1.0~\cite{li2024cosmicman}, as well as several high-quality internal datasets.
We perform careful data pre-processing and filtering, removing images with low-quality small faces, multiple faces, watermarks, or NSFW content. The data is pre-processed for training using aspect ratio bucketing~\cite{aspectratiobucketing}.
The total amount of single-person single-sample (SPSS) real data for stage-1 pre-training reaches $43$ million, which we consider sufficient for large-scale training of identity-preserved image generation models.
For stage-2 supervised fine-tuning, the total quantity of single-person-multiple-sample (SPMS) synthetic data is $2$ million. All data is generated by the stage-1 pretrained InfU model itself, equipped with useful off-the-shelf modules (see Section~\ref{subsec:method_mstraining}). High-quality synthetic data are also carefully processed and filtered to obtain image pairs with normal poses, high ID resemblance, and good aesthetics, ensuring their usefulness.
In addition, we observe that training the model with a mixture of captions from multiple sources, \eg, humans, small captioning models, and large vision-language models (VLMs), is beneficial. Besides the original captions in the datasets, we employ BLIP-2~\cite{blip2} and InternVL2~\cite{internvl} to obtain text captions from diverse sources for training.

\noindent
\textbf{Baselines.}
Since InfU is based on DiT (\eg, FLUX), we compare it with the most relevant and state-of-the-art DiT-based approach, PuLID-FLUX~\cite{pulid}. Other open-source efforts, including FLUX.1-dev IP-Adapters from InstantX~\cite{instantxfluxipa} and XLabs-AI~\cite{xfluxipa}, are not tailored for human faces. We select the one from InstantX as a representative baseline of this series for a more comprehensive comparison. Other conventional methods based on SDXL display much lower image quality due to the limitation of the base model (see Figure~\ref{fig:inspiration}) and are thus not fairly comparable. 

\noindent
\textbf{Evaluation.}
We conduct evaluations on a portrait benchmark created by GPT-4o~\cite{gpt4o}, comprising $200$ prompts and corresponding gender information. This benchmark covers a variety of cases, including different prompt lengths, face sizes, views, scenes, ages, races, complexities, \etc. We selected $15$ representative identity samples and paired their gender with all appropriate prompts, resulting in $1,497$ testing outputs for systematic evaluations.
We apply three representative and useful evaluation metrics, \ie, ID Loss~\cite{arcface}, CLIPScore~\cite{clipscore}, and PickScore~\cite{pickscore}.
ID Loss is defined as $1-\mathrm{CosSim}\left(\mathrm{ID}_{\text{gen}}, \mathrm{ID}_{\text{ref}}\right)$, where $\mathrm{CosSim}$ is cosine similarity, and $\mathrm{ID}_{\text{gen}}$ and $\mathrm{ID}_{\text{ref}}$ are the generated and reference identity images, respectively. A lower ID Loss means higher similarity. We follow the original papers to use CLIPScore and PickScore. A higher CLIPScore indicates better text-image alignment, and a higher PickScore signifies better image quality and aesthetics.

\begin{figure*}[t]
	\centering
	\vspace{-0.45cm}
	\includegraphics[width=0.99\linewidth]{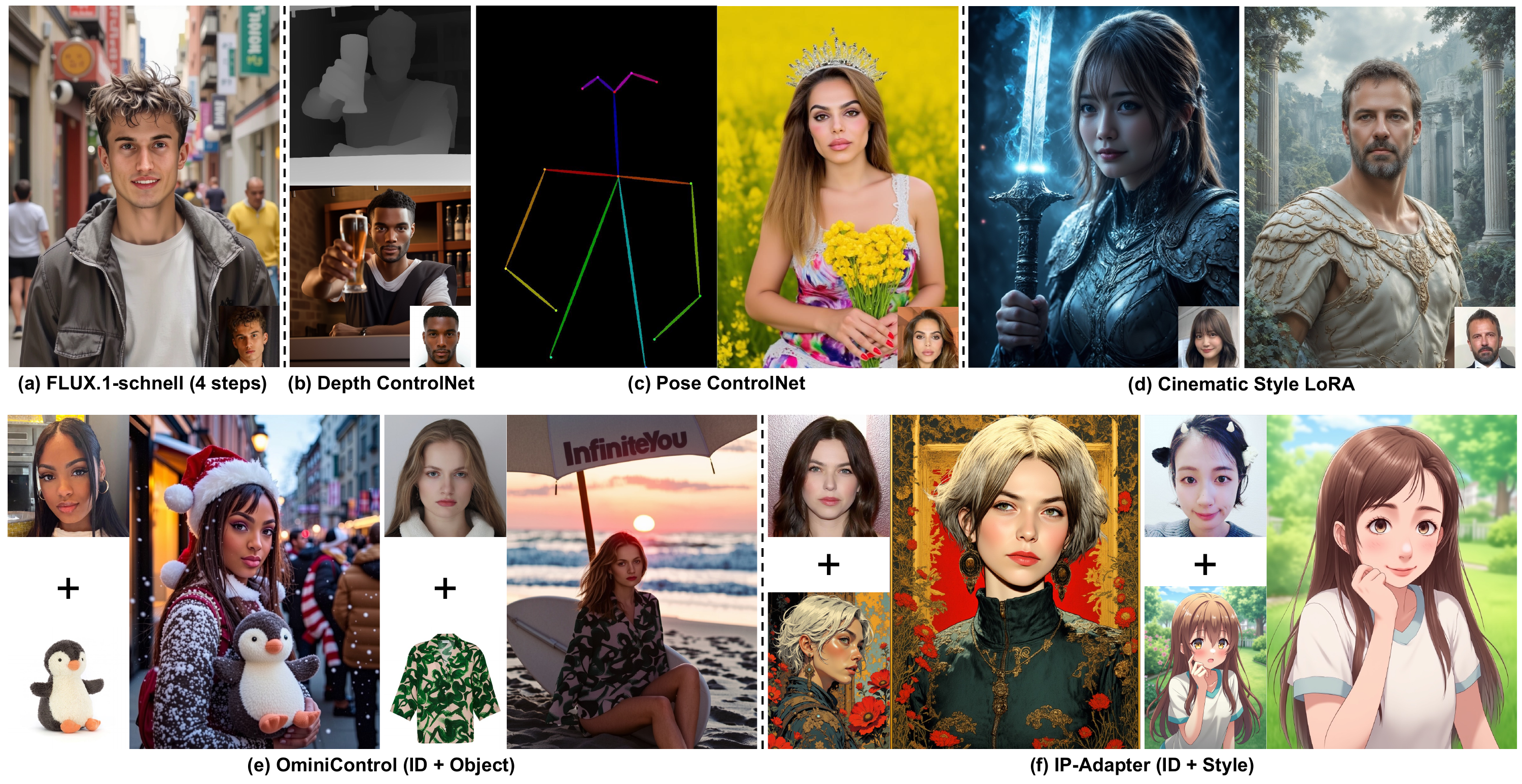}
	\vspace{-0.25cm}
	\caption{Desirable plug-and-play properties of InfU, compatible with many popular methods and plugins.}
	\label{fig:plug_and_play}
	\vspace{-0.3cm}
\end{figure*}

\subsection{Main Results}
\label{subsec:exp_results}

\noindent
\textbf{Qualitative comparison.}
The qualitative comparison results are shown in Figure~\ref{fig:qualitative_comparison}.
The identity similarity of the results generated by FLUX.1-dev IP-Adapter~(IPA)~\cite{instantxfluxipa} is inadequate. Besides, the text-image alignment and generation quality are inferior to other methods.
PuLID-FLUX~\cite{pulid} generates images with decent identity similarity. However, it suffers from poor text-image alignment (Column~$1$, $2$, $4$), and the image quality (\eg, bad hands in Column~$5$) and aesthetic appeal are degraded, indicating a large compromise in the generative capability of the base model. In addition, the face copy-paste issue is evident in the results generated by PuLID-FLUX (Column~$5$). In comparison, the proposed InfU outperforms the baselines across all dimensions. 

\noindent
\textbf{Quantitative comparison.}
The quantitative comparative results are shown in Table~\ref{tbl:quantitative_comparison}. 
Our method achieves the lowest ID Loss, indicating the best identity similarity. As mentioned, the existing release of FLUX.1-dev IPA~\cite{instantxfluxipa} is not tailored for human faces, resulting in much worse identity similarity than other methods.
In addition, our method obtains a significantly higher CLIPScore, demonstrating superior text-image alignment. Notably, the improvement in CLIPScore is substantial, reducing the gap to the upper-bound performance of FLUX.1-dev on our test set ($0.334$) by $66.7\%$.
Furthermore, our approach produces the best PickScore, suggesting that the overall image quality and generation aesthetics of InfU surpass all baselines.

\begin{table}[tb!]
\centering
\footnotesize
\caption{The ID Loss (lower is better) and CLIPScore (higher is better), and PickScore (higher is better) comparative results.}
\vspace{-0.3cm}
\begin{tabularx}{\linewidth}{c|*{3}{|Y}}
\Xhline{1pt}
Method& ID Loss$\downarrow$& CLIPScore$\uparrow$& PickScore$\uparrow$ \\
\cline{2-4}
\Xhline{0.6pt}
FLUX.1-dev IPA~\cite{instantxfluxipa}& 0.772& 0.243& 0.204 \\
PuLID-FLUX~\cite{pulid}& 0.225& 0.286& 0.212 \\
InfU (Ours)&  {\bf0.209}& {\bf0.318}& {\bf0.221} \\
\Xhline{1pt}
\end{tabularx}
\label{tbl:quantitative_comparison}
\vspace{-0.5cm}
\end{table}

\noindent
\textbf{User study.}
We conducted a user study on InfU and the most competitive baseline, PuLID-FLUX~\cite{pulid}. Participants were asked to evaluate $70$ sets of samples. The study included $16$ participants from diverse backgrounds (\eg, QA professionals, researchers, engineers, designers, \etc, from different countries) to reduce personal understanding bias. The best selection rate of our method reached $72.8\%$ in overall performance (in aspects of identity similarity, text-image alignment, image quality, and generation aesthetics), and PuLID-FLUX was $27.2\%$. This suggests that our results are significantly better \wrt human preference.

\noindent
\textbf{Plug-and-play properties.}
The proposed InfU method features a desirable plug-and-play design, compatible with many existing methods. 
It naturally supports base model replacement with any variants of FLUX.1-dev, such as FLUX.1-schnell~\cite{flux} for more efficient generation (e.g., in 4 steps, Figure~\ref{fig:plug_and_play}~(a)).
The compatibility with off-the-shelf ControlNets~\cite{controlnet} and LoRAs~\cite{lora} provides additional controllability and flexibility for customized tasks (Figure~\ref{fig:plug_and_play}~(b)(c)(d)).
Notably, our compatibility with OminiControl~\cite{ominicontrol} extends the potential of InfU for multi-concept personalization, such as interacted identity (ID) and object personalized generation (Figure~\ref{fig:plug_and_play}~(e)).
Although employing IP-Adapter (IPA)~\cite{ipa} with our method for identity injection is suboptimal (see Section~\ref{subsec:exp_ablation}), InfU is readily compatible with IPA for stylization of personalized images, producing decent results when injecting style references via IPA (Figure~\ref{fig:plug_and_play}~(f)).
Our plug-and-play feature can extend to even more approaches beyond those mentioned, providing valuable contributions to the broader community.

\subsection{Ablation Studies}
\label{subsec:exp_ablation}

\begin{table}[tb!]
\centering
\footnotesize
\caption{The ID Loss (lower is better) and CLIPScore (higher is better), and PickScore (higher is better) for ablation studies.} 
\vspace{-0.3cm}
\begin{tabularx}{\linewidth}{c|*{3}{|Y}}
\Xhline{1pt}
Method& ID Loss$\downarrow$& CLIPScore$\uparrow$& PickScore$\uparrow$ \\
\cline{2-4}
\Xhline{0.6pt}
w/o multi-stage training& {\bf0.172}& 0.292& 0.212 \\
w/o SPMS& 0.368& 0.303& 0.220 \\
w/ IPA& 0.180& 0.241& 0.199 \\
full method& 0.209& {\bf0.318}& {\bf0.221} \\
\Xhline{1pt}
\end{tabularx}
\label{tbl:ablation}
\vspace{-0.5cm}
\end{table}

We primarily conduct ablation studies on our core contributions of the multi-stage training strategy and the identity injection design. Since InfuseNet is indispensable, we highlight the importance of using InfuseNet solely, without incorporating IPA that could introduce negative impacts.

The results are shown in Table~\ref{tbl:ablation}.
Without stage-2 supervised fine-tuning (SFT), InfU can generate images with even higher identity similarity. However, text-image alignment degrades, and image quality and aesthetic appeal worsen. We deduce that the SPMS synthetic data introduces slightly more difficulty in learning identity, yet significantly improves other aspects.
Using single-person-single-sample (SPSS) synthetic data instead of SPMS in stage-2 SFT (w/o SPMS) results in a significant drop in identity similarity, as well as degraded text-image alignment and image quality.
We deduce that SPSS synthetic data may weaken the function of InfuseNet by directly learning a reconstruction of synthetic data rather than transforming reference real data into synthetic data. This may lead to fitting back to the base model's distribution without sufficient data diversity.
These results emphasize the importance of the multi-stage training strategy and the construction of the SPMS format. 
If we employ IPA together with InfuseNet for identity injection (distinct from stylization), text-image alignment, image quality, and aesthetics substantially deteriorate, despite a slight improvement in identity similarity (still worse than our stage-1 model). This underscores the non-optimality and negative effects of IPA.

\section{Conclusion}
\label{sec:conclusion}

We introduced InfU, a novel framework for identity-preserved image generation with advanced DiTs. InfU addresses key limitations of existing methods in identity similarity, text-image alignment, overall image quality, and generation aesthetics. Central to our framework is InfuseNet, which enhances identity preservation while maintaining generative capabilities.
The multi-stage training strategy further improves our overall performance.
Comprehensive experiments showed that InfU outperforms state-of-the-art baselines.
Moreover, InfU is plug-and-play and compatible with various methods, contributing significantly to the broader community.
InfU sets a new benchmark in this field, showcasing the immense potential of integrating DiTs for advanced personalized generation.
Future work may explore enhancements in scalability and efficiency, as well as expanding the application of InfU to other domains.

\noindent
\textbf{Limitations and societal impact.}
Despite promising results, the identity similarity and overall quality of InfU could be further improved. Potential solutions include additional model scaling and an enhanced InfuseNet design.
On another note, InfU may raise concerns about its potential to facilitate high-quality fake media synthesis. However, we believe that developing robust media forensics approaches can serve as effective safeguards.

\noindent
\textbf{Acknowledgments.}
We sincerely acknowledge the insightful discussions from Stathi Fotiadis, Min Jin Chong, Xiao Yang, Tiancheng Zhi, Jing Liu, and Xiaohui Shen.
We genuinely appreciate the help from Jincheng Liang and Lu Guo with our user study and qualitative evaluation.

\balance
{
    \small
    \bibliographystyle{ieeenat_fullname}
    \bibliography{main}
}

\clearpage

\end{document}